\pdfoutput=1
\documentclass[11pt]{article}

\usepackage[]{acl}

\usepackage{times}
\usepackage{latexsym}

\usepackage[T1]{fontenc}
\usepackage[utf8]{inputenc}

\usepackage{microtype}

\usepackage{amsmath}
\usepackage{amssymb}

\usepackage{multirow}
\usepackage{booktabs}
\usepackage{graphicx}
\usepackage{array}
\usepackage{color}
\usepackage{xcolor}
\usepackage{bbm}
\usepackage{subcaption}

\title{CLEAR: Improving Vision-Language Navigation with Cross-Lingual, Environment-Agnostic Representations}

\author{Jialu Li \quad \quad Hao Tan \quad \quad Mohit Bansal
 \\ 
   UNC Chapel Hill\\ 
   \texttt{\{jialuli, airsplay, mbansal\}@cs.unc.edu}
}

\begin{document}
\maketitle
\begin{abstract}
Vision-and-Language Navigation (VLN) tasks require an agent to navigate through the environment based on language instructions. In this paper, we aim to solve two key challenges in this task: utilizing multilingual instructions for improved instruction-path grounding and navigating through new environments that are unseen during training. To address these challenges, we propose `\textbf{CLEAR}: \textbf{C}ross-\textbf{L}ingual and \textbf{E}nvironment-\textbf{A}gnostic \textbf{R}epresentations'. First, our agent learns a shared and visually-aligned cross-lingual language representation for the three languages (English, Hindi and Telugu) in the Room-Across-Room dataset. Our language representation learning is guided by text pairs that are aligned by visual information. Second, our agent learns an environment-agnostic visual representation by maximizing the similarity between semantically-aligned image pairs (with constraints on object-matching) from different environments. Our environment agnostic visual representation can mitigate the environment bias induced by low-level visual information. Empirically, on the Room-Across-Room dataset, we show that our multilingual agent gets large improvements in all metrics over the strong baseline model when generalizing to unseen environments with the cross-lingual language representation and the environment-agnostic visual representation.
Furthermore, we show that our learned language and visual representations can be successfully transferred to the Room-to-Room and Cooperative Vision-and-Dialogue Navigation task, and present detailed qualitative and quantitative generalization and grounding analysis.\footnote{Code and model are available at \url{https://github.com/jialuli-luka/CLEAR}.}
\end{abstract}

\section{Introduction}

\begin{figure}[t]
\begin{center}
\includegraphics[width=0.99\linewidth]{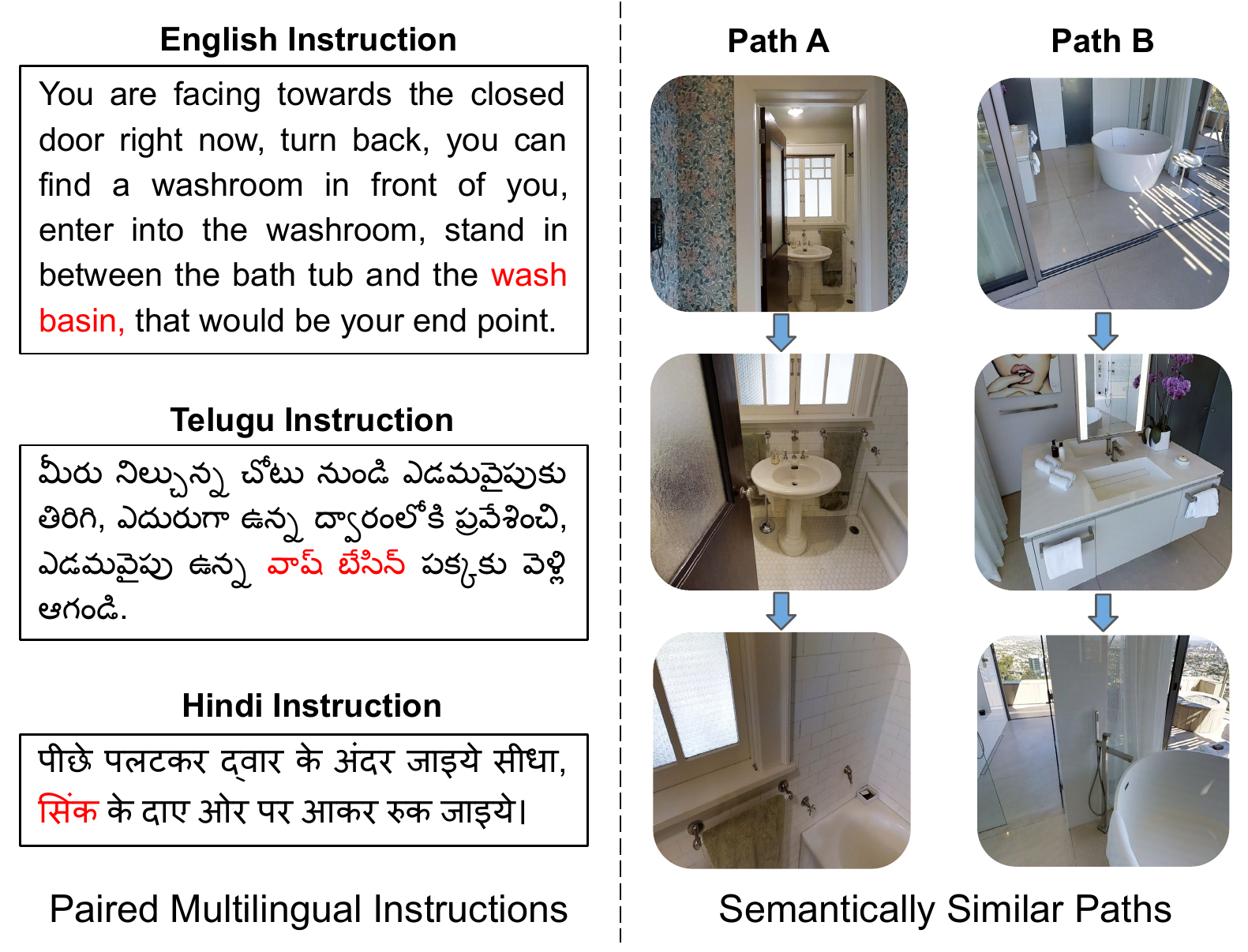}
\end{center}
   \vspace{-8pt}
  \caption{Motivation for cross-lingual and environment-agnostic visual representations: The English instruction, Telugu instruction, Hindi instruction on the left all correspond to the same path -- Path A. The words in red correspond to the same visual object ``wash basin". Path A and Path B are similar paths (i.e., the instruction for these two paths are semantically similar) if they contain same object(s) in different environments.
  }
  \vspace{-8pt}
\label{figure1}
\end{figure}

The Vision-and-Language Navigation task requires an agent to navigate through the environment based on language instructions. This task has two unsolved challenges. 
First, directly introducing pre-trained linguistic and visual representations into these agents suffers from domain shift (i.e., pre-trained linguistic and visual representation might not generalize to VLN task)~\cite{huang2019transferable}. Learning the instruction representation while also learning how to navigate based on the instruction is even more challenging for a multilingual agent, since more language variance is injected via multilingual instructions. At the same time, it also poses the important question that whether we can utilize multilingual instructions to learn a better cross-lingual representation and improve instruction-path grounding and referencing. 
Second, previous works \citep{fried2018speaker, wang2019reinforced, landi2019perceive, wang2020active,  Huang2019MultimodalDM, ma2019selfmonitoring, majumdar2020improving, Qi2020ObjectandActionAM} on vision-language navigation have seen that agents tend to perform substantially worse in environments that are unseen during training, indicating the lack of generalizability of the navigation agent. In this paper, we propose to address these two challenges via cross-lingual and environment agnostic representations.

Although some initial progress~\cite{huang2019transferable,  majumdar2020improving, hong2020recurrent, chen2021history} has been made towards introducing pre-trained linguistic representations into vision-language navigation agents, how to understand and utilize paired multilingual instructions to transfer the pre-trained linguistic representation to multilingual navigation agents still remains unexplored. We argue that for a multilingual agent, the linguistic representation can capture more visual concepts from learning the similarity between paired multilingual instructions. As shown in Figure \ref{figure1}, though the three instructions shown here are in different languages and vary in length and level of detail\footnote{We translate Telugu instruction and Hindi instruction into English instruction with Google Translation for reference here (the translated instructions are not used in representation learning or navigation learning). Telugu: Return to the left from where you are standing, enter the door on the opposite side, and go to the side of the wash basin on the left and wait. Hindi: Turn back and go inside the door directly, come to the right side of the sink and stop.}, all of them correspond to the same path -- Path A. Hence, by learning the similarity between these paired instructions, the cross-lingual language representation of the same visual concept mentioned in these paired instructions (e.g., the red words correspond to the same visual object ``wash basin") will be close to each other, making it easier for the agent to comprehend. Furthermore, the cross-lingual language representation will benefit from the complementary information from instructions in different languages since they elicit more references to visible entities. For example, in Figure \ref{figure1}, the target room environment ``washroom" is only mentioned in English instructions. Hindi and Telugu instructions could benefit from learning the connection between ``washroom" and ``wash basin" through learning from the English instruction.

Moreover, many methods have been proposed to encourage agents' generalization to unseen environments during training~\cite{tan-etal-2019-learning, wang2020environment, fu2020counterfactual, ZhangTB20}. \citet{ZhangTB20} has shown that it is the low-level appearance information that causes the environment bias. To mitigate this bias, previous works only consider one single environment when learning the visual representation for a given path. We instead learn an environment-agnostic visual representation by exploring the connections between multiple environments. For the example shown in Figure \ref{figure1}, Path A and Path B are two semantically aligned paths in different environments. In both cases, the agent needs to head into the washroom and stop beside the wash basin. Learning the relationship between these paired paths helps the agent comprehend concepts like ``bath tub", and not be distracted by the low-level appearance of the objects in unseen environments.

\begin{figure*}[t]
\begin{center}
\includegraphics[width=0.99\linewidth]{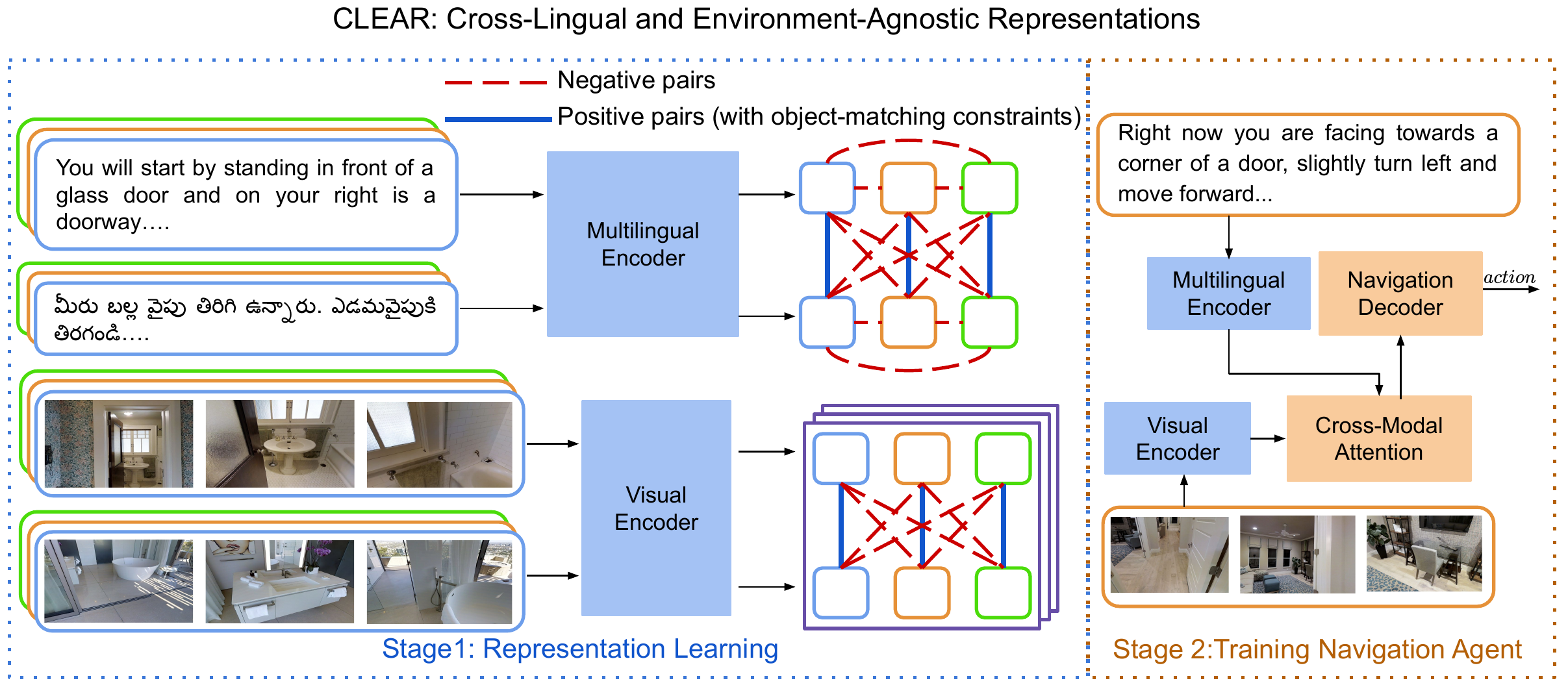}
\end{center}
\vspace{-8pt}
  \caption{Overview of CLEAR (Cross-Lingual and Environment-Agnostic Representations). Left: the agent learns a cross-lingual language representation and an environment-agnostic visual representation via maximizing the similarity between positive pairs (connected with blue line) and minimizing the similarity between negative pairs (connected with red dashed line). For simplicity, we use 3 as batch size when illustrating the positive pairs and negative pairs. Right: then the agent is trained on the vision-and-language navigation task based on these learned representations.}
  \vspace{-5pt}
\label{figure2}
\end{figure*}

Overall, in this paper, we propose `\textbf{CLEAR}: \textbf{C}ross-\textbf{L}ingual and \textbf{E}nvironment-\textbf{A}gnostic \textbf{R}epresentations' to address the two challenges above. First, we define a visually-aligned instruction pair as two instructions that correspond to the same navigation path. Given the instruction pairs, we transfer the pre-trained multilingual BERT~\cite{devlin-etal-2019-bert} to the Vision-Language Navigation task by encouraging these paired instructions to be embedded close to each other. Second, we identify semantically-aligned path pairs based on the similarity between instructions. Intuitively, if the similarity between the two instructions is high, then their corresponding navigation path will be semantically similar (i.e., mentioning the same objects like ``wash basin"). We further filter out image pairs (a pair of paths will contain multiple image pairs) that do not contain the same objects for higher path pair similarity. Then, we train an environment agnostic visual representation that learns the connection between these semantically-aligned path pairs.

We conduct experiments on the Room-Across-Room (RxR) dataset~\cite{ku-etal-2020-room}, which contains instructions in three languages (English, Hindi, and Telugu). Empirical results show that our proposed representations significantly improve the performance over the mono-lingual model~\cite{shen2021much} by 2.59\% in nDTW score on RxR test leaderboard. 
We further show that our CLEAR approach outperforms our baseline that utilizes ResNet~\cite{he2016deep} to extract image features by 5.3\% in success rate and 4.3\% in nDTW score (and it also outperforms a stronger baseline that utilizes the recent CLIP~\cite{radford2021learning} method to extract image features). 
Moreover, our CLEAR approach shows better generalizability when transferred to Room-to-Room (R2R) dataset~\cite{anderson2018vision} and Cooperative Vision-and-Dialogue Navigation dataset~\cite{thomason:arxiv19}, and adapted to other SotA VLN Agent~\cite{chen2021history}. 
We also demonstrate the advantage of optimizing similarity between all the three languages in RxR dataset for language representation learning and the effectiveness of the way we generate positive path pairs for visual representation learning.
Lastly, we demonstrate that our cross-lingual language representation captures visual semantics underlying the instructions, and our environment-agnostic visual representation generalizes better to the unseen environment with both qualitative and quantitative analysis.

\section{Related Work}
\noindent\textbf{Vision-and-language navigation.}
Vision-and-Language Navigation (VLN) requires an agent to find the routes to the desired target based on instructions~\cite{jain-etal-2019-stay, thomason2020vision, nguyen-daume-iii-2019-help, qi2020reverie, chen2019touchdown, krantz2020beyond}.
Specifically, there are two key challenges in VLN: grounding the natural language instruction to visual environments and generalizing to unseen environments. 
To address the first challenge, one line of research in VLN utilizes carefully designed cross-modal attention modules~\cite{wang2018look, wang2019reinforced, tan-etal-2019-learning, landi2019perceive, Xia2020MultiViewLF, wang2020soft, wang2020active, DBLP:conf/acl/ZhuHCDJIS20, li-etal-2021-improving, zhu2021soon, an2021neighbor, kim2021ndh}, progress monitor modules~\cite{ma2019regretful, ma2019selfmonitoring, ke2019tactical}, and object-action aware modules~\cite{Qi2020ObjectandActionAM}. Another line of research improves vision and language co-grounding by improving vision and language representations with pre-training techniques~\cite{Li2019RobustNW, huang2019transferable, hao2020towards, majumdar2020improving, hong2020recurrent}. \citet{Li2019RobustNW} directly adopts pre-trained BERT for encoding instructions, \citet{hao2020towards} and \citet{hong2020recurrent} learn from a large amount of image-text-action triplets, \citet{majumdar2020improving} learns from large amount of text-image pairs from the web, and \citet{huang2019transferable} transfers language and visual representation to in-domain representation with auxiliary tasks. Different from them, we utilize visually-aligned multilingual instructions to learn a cross-lingual language representation that inherently captures the visual semantics underlying the instruction.

Multiple methods have been proposed to encourage generalization to unseen environments during training \cite{ZhangTB20, tan-etal-2019-learning, wang2020environment, fu2020counterfactual, li2022envedit}. \citet{ZhangTB20} demonstrates that it is the low-level appearance information that causes the large performance gap between seen and unseen environments. \citet{tan-etal-2019-learning} proposes to use environment dropout on visual features to create new environments and \citet{fu2020counterfactual} utilizes adversarial path sampling to encourage generalization. However, both of these methods rely on a speaker module to generate synthetic training data and can be considered as data augmentation methods, which are complementary to our proposed environment-agnostic visual representation. The closest work to ours is \citet{wang2020environment}, where they propose to pair an environment classifier with a gradient reversal layer to learn an environment-agnostic representation. However, they only consider one single environment when learning the visual representation for a given path (i.e., given one path and predict its environment). In our environment-agnostic representation learning, we explore the connections between multiple environments (i.e., maximize the similarity between paths from different environments). 

\noindent\textbf{Vision-and-language with multilinguality.} 
There has been growing interest in combining vision and language for tasks such as visual-guided machine translation \cite{sigurdsson2020visual, suris2020globetrotter, huang2020unsupervised}, multilingual visual question answering \cite{gao2015you, gupta2020unified,shimizu2018visual}, multilingual image captioning \cite{gu2018unpaired, lan2017fluency}, multilingual video captioning \cite{wang2019vatex}, and multilingual image-sentence retrieval \cite{kim2020mule, burns2020learning}. In this paper, we work on multilingual vision-and-language navigation. We use vision (i.e., navigation path) as a bridge between multilingual instructions and learn a cross-lingual representation that captures visual concepts. Moreover, our method uses language as a bridge between different visual environments to learn an environment-agnostic visual representation.

\section{Method}

In this section, we present our CLEAR method that learns cross-lingual language representations and environment-agnostic visual representations. Given these learned language and visual representations, we then train the agent on the vision-and-language navigation task with imitation learning and reinforcement learning.
The overall representation learning and navigation agent training processes are illustrated in Figure~\ref{figure2}. 
We next describe our representation learning methods in Sec.~\ref{section:language} and Sec.~\ref{section:visual}. The navigation model \cite{tan-etal-2019-learning} and training process are detailed in Appendix.

\subsection{Language Representation Learning} \label{section:language}

The goal of our language representation learning approach is to learn a cross-lingual language representation that can mitigate the natural ambiguity and variance in multilingual instructions and improve the path-instruction alignment by capturing the shared and salient visual concepts underlying the instructions. We define visually-aligned instruction pairs as instructions that correspond to the same navigation path. 
Since these instruction pairs refer to the same navigation path, the visual concepts underlying these instructions (e.g., visual objects mentioned in the instruction) are shared. 
Thus, we could train the language representation to emphasize these visual concepts by learning the connection between these visually-aligned instruction pairs. 

For each navigation path, the Room-Across-Room (RxR) dataset \cite{ku-etal-2020-room} provides 9 corresponding language instructions in 3 languages (English, Hindi, and Telugu). 
During training, for each navigation path, we randomly sample two instructions out of the nine corresponding instructions as the visually-aligned instruction pair.
The two instructions can be in different languages, which helps the agent learn a cross-lingual language representation. 
Exclusively learning connections between instructions in the same language will lose crucial information across languages, and we quantitatively illustrate this result in Sec.~\ref{section: lang_quality}. 

Given the instruction $\{w_{i}\}_{i=0}^{m}$ with $m$ words, we use feature of the $\mathrm{[CLS]}$ token (i.e., $w_0$) in the pre-trained multilingual BERT~\cite{devlin-etal-2019-bert} outputs as the sentence representation $\widetilde{w}$:
\begin{align}
    \{\widehat{w}_i\}_{i=0}^m &= \mathrm{m\mbox{-}BERT}(\{w_{i}\}_{i=0}^{m}) \\
    \widetilde{w} &= \widehat{w}_0  \label{eqn:lang}
\end{align}
In a batch of size $N$, we have $N$ positive pairs of instructions with representations $(\widetilde{w}_j, \widetilde{u}_j)_{j=1}^N$ from Eqn.~\ref{eqn:lang}. Each positive pair is matched with $2(N - 1)$ negatives in the batch (i.e., $\{\widetilde{w}_k\}_{k\neq i}$ and $\{\widetilde{u}_k\}_{k\neq j}$). 
Our goal is to learn a representation that maps instructions for the same path closer to each other in the representation space, regardless of the language and the natural variance in human-generated instructions.
We learn the representation by optimizing a contrastive loss: 
\begin{align}
\label{equ:lloss}
    L_{lang} &= -\sum_{i=1}^N{\log\frac{\exp(\alpha_{i,i} / \tau)}{\sum_{k=1}^{2N}\mathbbm{1}_{k\neq{i}}\exp(\alpha_{i,k} / \tau)}} \\ 
    \alpha_{i,j} &= \frac{\widetilde{w}_i^T \widetilde{u}_j}{\|\widetilde{w}_i\| 	\|\widetilde{u}_j\|}
\end{align}
where $\alpha_{i,j}$ is the similarity between the instruction $\widetilde{w}_i$ and $\widetilde{u}_j$, and $\tau$ is the temperature hyperparameter. 

\subsection{Visual Representation Learning} \label{section:visual}
Our goal in visual representation learning is to learn an environment-agnostic visual representation that can mitigate the environment bias caused by objects' low-level appearance, such that it could generalize better to unseen environments. Intuitively, the agent would learn the general concept of objects instead of the low-level appearance if the agent can identify the same objects in two images in different environments. Thus, we train the agent to learn the connected visual semantics between the semantically-aligned navigation paths (i.e., paths that mention the same objects or mention similar actions in different environments). 

\noindent\textbf{Identifying semantically-aligned path pairs.} 
Although the appearance of the path varies a lot in different environments, the instructions that describe similar paths are more consistent across environments. Based on this intuition, we use language as the bridge between paths in multiple visual environments.
Specifically, we propose to use instruction similarity as a direct measurement of how semantically similar two paths are. For each instruction-path pairs $(I,P)$ given in the Room-Across-Room (RxR) dataset, we first represent each instruction $I$ as in Eqn. \ref{eqn:lang}. Then, we compute the cosine similarity between the representation of instruction $I$ and all the other instructions in the training set. We pick the instruction $\widehat{I}$ that is most similar to $I$  and also constraints that $\widehat{I}$'s corresponding path $\widehat{P}$ has the same path length as $P$. Thus, we group $P$ and $\widehat{P}$ as the semantically-similar path pair. 

\noindent\textbf{Constraint on object-matching.}
In a batch of size N, we have N positive semantically-aligned path pairs $(P_{k}, Q_{k})_{k=1}^{N}$. We represent the positive path pair $(P_{k}, Q_{k})$ as sequences of panoramic views $(\{p_{k,t}\}_{t=1}^{L_{k}}, \{q_{k,t}\}_{t=1}^{L_{k}})$ with length $L_k$. Since paths might not be fully aligned (i.e., correspondence between image pairs $\{p_{k,t}\}$ and $\{q_{k,t}\}$ might not hold), we use object-matching to filter out image pairs that don't contain the same objects. Specifically, we use Mask-RCNN \cite{he2017mask} model trained on LVIS dataset \cite{gupta2019lvis} in detectron2 \cite{wu2019detectron2} to detect objects in the 36 discretized views of the panoramic view. We filter out object classes that appear less than 1\% of the time in all panoramic views. 27 object classes left, including objects like `cabinet', `chair', and `sofa'. All object classes can be found in Appendix. During training, we randomly sample 10 out of 27 object classes in each iteration and filter out image pairs that don't contain same objects of the sampled 10 object classes. Our object-matching constraint ensures that the corresponding image pairs $\{p_{k,t}\}$ and $\{q_{k,t}\}$ also have a high semantic similarity.

\noindent\textbf{Visual encoder.}
The panoramic view of time step $t$ is discretized into 36 single views $\{o_{t,i}\}_{i=1}^{36}$. We encode the visual representation for each view as:
\begin{align}
    \widehat{o}_{t,i} &= \mathrm{pre}\mbox{-}\mathrm{trained\ model}(o_{t,i}) \\
    v_{t,i} &= W_{v1}\mathrm{ReLU}(W_{v2}\widehat{o}_{t,i}) \\
    \widehat{v}_{t,i} &= \mathrm{LayerNorm}(v_{t,i} + \widehat{o}_{t,i})
\end{align}
We first encode images with pre-trained vision models. Then the encoded view features are passed through two fully-connected layers with ReLU as activation function. Layer normalization and residual connection are applied on top of the fully-connected layer.

\noindent\textbf{Learning visual representation:} 
Given the N positive semantically-aligned path pairs $(P_{k}, Q_{k})_{k=1}^{N}$, at each time step $t$, we have $N_p$ panoramic views (computed as the average of 36 single views as in Eqn.~\ref{equ:panoview}) that have a positive pair (i.e., the paired view contain at least one same object). For each view $p_{k,t}$ that has a positive pair, the visual encoder is trained to predict which of the $N$ possible panoramic views $\{q_{k,t}\}_{k=1}^N$ contain similar semantic information. Specifically, we train the visual encoder to maximize the cosine similarity of the $N_p$ positive image pairs in the batch while minimizing the cosine similarity of the $N*N_p-N_p$ negative image pairs (i.e. each view has $N-1$ negatives). We optimize the 
contrastive loss as:
\begin{align}
\label{equ:vloss}
    L_{visual} &= -\sum_{k=1}^{N_p}\sum_{t=1}^{L_k}\log(\mathrm{Softmax}_k(\beta_{k,t}/\tau)) \\ 
    \beta_{k,t} &= \frac{p_{k,t}^Tq_{k,t}}{\|p_{k,t}\| 	\|q_{k,t}\|} 
\end{align}
where $\beta_{k,t}$ is the similarity between positive panoramic view pair $p_{k,t}$ and $q_{k,t}$, and $\tau$ is the temperature hyperparameter. We compute the panoramic view representation as the average of 36 single views:
\begin{align}
\label{equ:panoview}
    p_{k,t} &= \frac{1}{36}\sum_{i=1}^{36}\widehat{v}_{p,k,t,i}
\end{align}
where $\widehat{v}_{p,k,t,i}$ is the output representation from the visual encoder. $q_{k,t}$ is computed similarly.

\subsection{Learning}
Our CLEAR agent has two stages of learning: representation learning and navigation learning. 

In the representation learning stage, we train the multilingual encoder and visual encoder by optimizing the contrastive loss $L_{lang}$ in Eqn.~\ref{equ:lloss} and $L_{visual}$ in Eqn.~\ref{equ:vloss} respectively. 
The representation learning process transfers the language representation to domain-specific language representation and adapts the visual representation to learn the correlation underlying the navigation environments. 

In the navigation learning stage, we use a mixture of imitation learning and reinforcement learning to train the agent on the navigation task as in \citet{tan-etal-2019-learning}. Details can be found in Appendix.

\section{Experimental Setup}
\subsection{Dataset}
We evaluate our agent on the Room-Across-Room (RxR) dataset \cite{ku-etal-2020-room}. The dataset is split into training set, seen and unseen validation set, and test set. In the unseen validation set and test set, the environments are not appeared in training set. Thus the performance on these two sets shows the model's generalizability to new environments. More details can be found in Appendix.

\begin{table}[t]
    \centering
    \begin{tabular}{cccccc}
    \hline 
       \textbf{Models}   &  \textbf{SR$\uparrow$} & \textbf{SPL$\uparrow$} & \textbf{NDTW$\uparrow$} & \textbf{SDTW$\uparrow$} \\ \hline
        \textbf{RxR}  & 20.98 & 18.55 & 36.81 & 16.88  \\
    \textbf{CLIP} & 38.34 & 35.17 & 51.10 & 32.42 \\
    \textbf{Our}  &  40.29 & 36.57 & 53.69 & 34.86 \\ 
    \hline
    \end{tabular}
    \caption{Test leaderboard results under single run setup. RxR is the mono-lingual baseline in \citet{ku-etal-2020-room}, CLIP is the mono-lingual agent in \citet{shen2021much}}
    \label{table1}
    \vspace{-10pt}
\end{table}

\begin{table*}[t]
\begin{small}
\centering
\begin{tabular}{p{0.07\columnwidth}>{\centering\arraybackslash}p{0.060\columnwidth}>{\centering\arraybackslash}p{0.060\columnwidth}>{\centering\arraybackslash}p{0.060\columnwidth}>{\centering\arraybackslash}p{0.060\columnwidth}>{\centering\arraybackslash}p{0.060\columnwidth}>{\centering\arraybackslash}p{0.060\columnwidth}>{\centering\arraybackslash}p{0.060\columnwidth}>{\centering\arraybackslash}p{0.060\columnwidth}>{\centering\arraybackslash}p{0.060\columnwidth}>{\centering\arraybackslash}p{0.060\columnwidth}>{\centering\arraybackslash}p{0.060\columnwidth}>{\centering\arraybackslash}p{0.060\columnwidth}>{\centering\arraybackslash}p{0.060\columnwidth}>{\centering\arraybackslash}p{0.060\columnwidth}>{\centering\arraybackslash}p{0.060\columnwidth}>{\centering\arraybackslash}p{0.060\columnwidth}}
\hline 
\multicolumn{1}{c}{\textbf{Models}} & \multicolumn{4}{c}{\textbf{SR$\uparrow$}} & \multicolumn{4}{c}{\textbf{SPL$\uparrow$}} & \multicolumn{4}{c}{\textbf{NDTW$\uparrow$}} & \multicolumn{4}{c}{\textbf{SDTW$\uparrow$}} \\ \hline
 & avg & en & hi & te & avg & en & hi & te & avg & en & hi & te & avg & en & hi & te \\ \hline
\textbf{RxR} & 22.8 & 22.2 & 23.0 & 23.1 & 20.4 & 19.8 & 20.7 & 20.7 & 38.9 & 38.6 & 39.2 & 38.8 & 18.2 & 17.8 & 18.3 & 18.4 \\  \hline
\textbf{ResNet} & 35.1 & 35.4 & 36.4 & 33.4 & 31.6 & 31.6 & 33.0 & 30.4 & 51.1 & 50.7 & 52.3 & 50.3 & 30.1 & 30.1 & 31.4 & 28.7 \\
\textbf{+text} & 36.0 & 36.1 & 37.6 & 34.3 & 31.7 & 31.7 & 33.2 & 30.3 & 52.0 & 52.3 & 53.4 & 50.2 & 30.5 & 30.5 & 32.0 & 29.1 \\ 
\textbf{+visual} & 35.6  & 35.8 & 36.9 & 33.9 & 32.5 & 32.6 & 33.9 & 31.0 & 53.7 & 53.6 & 55.1 & 52.5 & 30.5 & 30.5 & 31.7 & 29.1 \\ 
\textbf{+both} & 40.4 & 41.5 &42.2 & 37.6 & 36.5 & 36.7 & 38.5 &  34.3 & 55.4 & 54.4 & 57.8 & 54.1 &  34.6 & 35.1 & 36.4 & 32.2 \\ \hline
\textbf{CLIP} & 41.7 & 42.5 & 44.0 & 38.6 & 37.1 & 37.2 & 39.2 & 34.8 & 55.8 & 55.6 & 57.3 & 54.5 & 35.6 & 36.3 & 37.6 & 33.3 \\
\textbf{+both} & \textbf{44.4} & \textbf{46.0} & \textbf{46.0} & \textbf{41.1} & \textbf{39.3} & \textbf{40.1}  & \textbf{41.0} & \textbf{36.9} & \textbf{57.0} & \textbf{57.2} & \textbf{58.1} & \textbf{55.7} & \textbf{37.8} & \textbf{38.7} & \textbf{39.3} & \textbf{35.3} \\ \hline
\end{tabular}\\

\caption{Ablation study of our model with ResNet features and CLIP features on validation unseen sets. `avg' is the agent's average performance on English, Hindi, and Telugu instructions.}
\vspace{-10pt}
\label{table2}
\end{small}
\end{table*}

\subsection{Evaluation Metrics}
To evaluate the performance of our model, we follow the metrics used in the Room-Across-Room paper \cite{ku-etal-2020-room} (details in Appendix): Success Rate (SR), Success rate weighted by Path Length (SPL) \cite{anderson2018evaluation}, normalized Dynamic Time Warping (nDTW) \cite{magalhaes2019effective}, and success rate weighted by Dynamic Time Warping (sDTW) \cite{magalhaes2019effective}. nDTW and sDTW are the main metrics for RxR, and SR and SPL are the main metrics for R2R.

\subsection{Implementation Details}
In our experiments, we learn the shared cross-lingual representation based on cased multilingual $\text{BERT}_\text{BASE}$. For the pre-trained vision model, we compare performance between image features extracted from ImageNet-pre-trained \cite{russakovsky2015imagenet} ResNet-152 \cite{he2016deep} and CLIP-pre-trained \cite{radford2021learning} vision transformer (ViT-B/32) \cite{dosovitskiy2010image} (abbreviated as `CLIP feature' later). More details about representation learning and navigation training can be found in Appendix.

\section{Results}

\subsection{Test Set Results}
We compare our final agent model with results on the Room-Across-Room (RxR) leaderboard. Our agent is a multilingual model that learns three languages in the same model. Compared with monolingual agents that learn instructions in three languages separately, a multilingual agent performs worse due to high-resource languages degradation \cite{ku-etal-2020-room, aharoni2019massively, pratap2020massively}. Our agent is tested under the single-run setup. 
In the single-run setting, the agent only navigates once and does not pre-explore the test environment. 
As shown in Table~\ref{table1}, our CLEAR model with CLIP features is 16.88\% higher in nDTW score than the baseline mono-lingual model \cite{ku-etal-2020-room} (`RxR') that utilizes ResNet features and other base navigation model. Furthermore, our model is 2.59\% higher in nDTW score than the mono-lingual model \cite{shen2021much} (`CLIP') that utilizes CLIP features and the same base navigation model as ours.

\subsection{Ablation Results}
We demonstrate the effectiveness of our learned visual and language representations with ablation studies. 
The baseline model (annotated as `ResNet' in Table~\ref{table2}) uses multilingual BERT and pre-trained ResNet to encode instructions and images without the representation learning stage. 
Our CLEAR-ResNet (`ResNet+both' in Table \ref{table2}) outperforms its baseline models in all evaluation metrics on average. 
Specifically, it improves the baseline model by 5.3\% in success rate (SR) and 4.3\% in nDTW score on average over three languages. 
These results demonstrate that our CLEAR agent is not only more capable of reaching the target, but also follows the ground-truth path better.

We then show that both the cross-lingual language representation and environment-agnostic visual representation contribute to the overall improvement. 
When the cross-lingual language representation is added (`+text'), we see consistent improvement on the averaged metrics and observe that Hindi benefits most from the cross-lingual language representation. When adding the environment-agnostic visual representation (`+visual'), the nDTW score improves by 2.6\%. These improvements validate the effectiveness of our learned language and visual representations.

Moreover, we show that our CLEAR approach could generalize to other pre-trained visual features. We implement another model (annotated as `CLIP' in Table~\ref{table2}) that uses CLIP to encode images, which is a stronger baseline compared with the ResNet baseline (`ResNet' in Table~\ref{table2}). Our CLEAR-CLIP model (`CLIP+both' in Table \ref{table2}) also shows 2.7\% improvement in success rate (SR) and 1.2\% improvement in nDTW score on average over three languages. This demonstrates the effectiveness of our CLEAR approach over different pre-trained visual features.

\begin{figure*}[t]
\begin{center}
\includegraphics[width=0.99\linewidth]{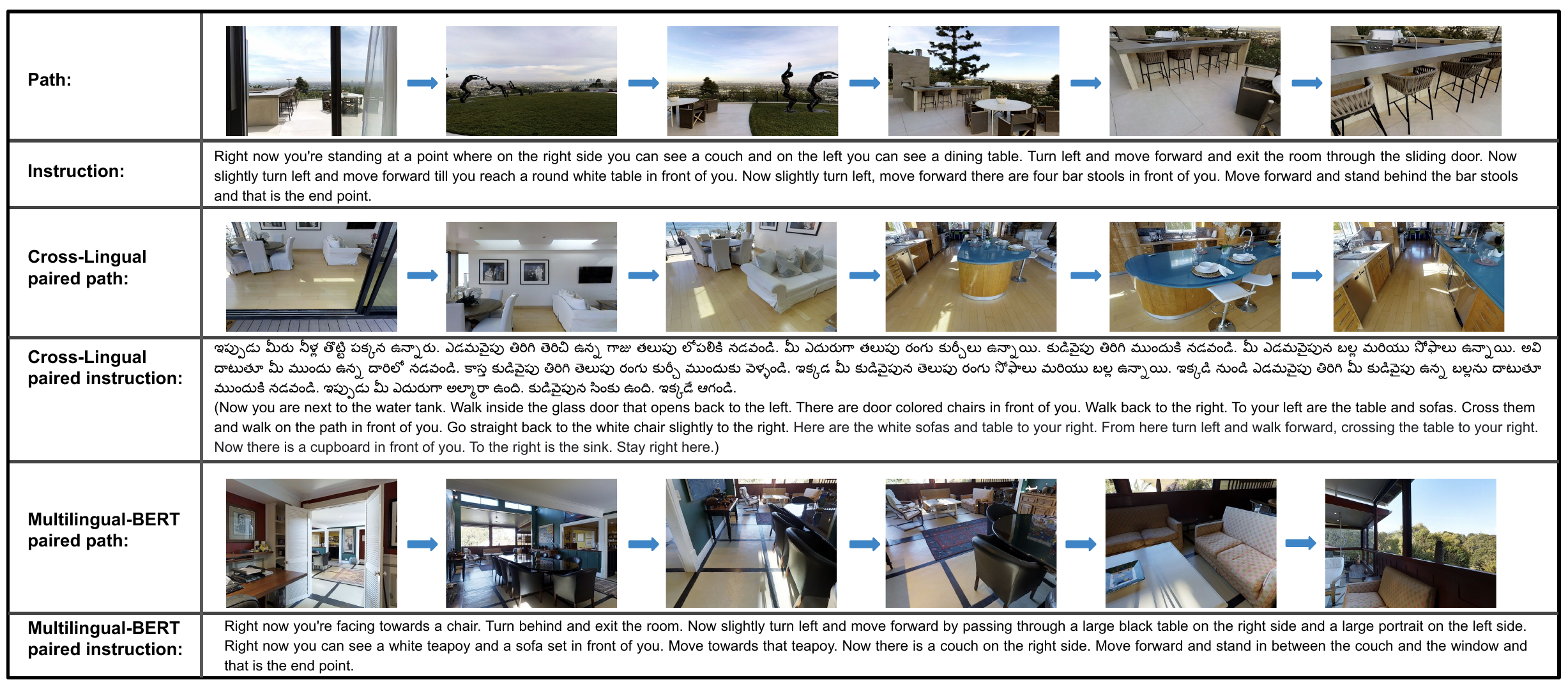}
\end{center}
\vspace{-10pt}
  \caption{Comparison of the most similar instruction picked with cross-lingual representation and multilingual-BERT. Our cross-lingual picked instruction mentions more visual object as in the query instruction. Besides, the path corresponding to the cross-lingual picked instruction contains more accurate visual objects as in the query path.}
  \vspace{-8pt}
\label{figure3}
\end{figure*}

\begin{table}
\centering
\resizebox{0.9\columnwidth}{!}{
    \begin{tabular}{cccccc}
    \hline 
       \textbf{Methods}  &  \textbf{SR$\uparrow$} & \textbf{SPL$\uparrow$} & \textbf{NDTW$\uparrow$} & \textbf{SDTW$\uparrow$} \\ \hline
        \textbf{m-BERT}  & 35.1 & 31.6 & 51.1 & 30.1 \\
        \textbf{Mono}  & 32.9 & 30.4 & 51.4 & 28.0 \\
    \textbf{Multi}   & 36.0 & 31.7 & 52.0 & 30.5 \\  \hline 
    \end{tabular}}
    \caption{Comparison between language representation trained with mono-lingual instruction pairs (`Mono') and multilingual instruction pairs (`Multi') on validation unseen sets. `m-BERT' is the method that uses original multilingual BERT as language representation.}
    \label{table4}
    \vspace{-10pt}
\end{table}

\section{Analysis}

We demonstrate the effectiveness of our cross-lingual representation in Sec.~\ref{section: lang_quality} and Sec.~\ref{sec:3lang}. We then show that our learned environment agnostic visual representation can decrease the gap between seen and unseen environments in Sec.~\ref{sec:gap}. Next, we also compare different contrastive learning approaches in Sec.~\ref{section:simcse}. Lastly, we show our agents' generalization to other datasets (Sec.~\ref{sec:generalization}) and another VLN agent (Sec.~\ref{sec:vln_agent}). More analysis on effectiveness of object matching constraints, helping decrease variance across environments, learning better multi-modal alignment and better word representation, correspondence between instruction similarity and path pair alignment can be found in Appendix.

\subsection{Effectiveness of Cross-Lingual Representations} \label{section: lang_quality}
In this section, we show the effectiveness of our language representation learning method described in Sec.~\ref{section:language}.
We first show the effectiveness of using paired multilingual instructions instead of monolingual instructions in the language representation learning stage. 
Then, we show that our learned cross-lingual language representation captures the visual concepts behind the instruction better than the original multilingual BERT representation.

\noindent\textbf{Multilingual vs. monolingual.}
To show that the multilingual instruction pairs are crucial for our cross-lingual language representation learning, we experiment with fine-tuning multilingual BERT with instruction pairs in same language only (`Mono' in Table~\ref{table4}). We observe that compared with the agent with cross-lingual representation (`Multi'), the success rate decreases by 3.1\% and sDTW score decreases by 2.5\%. Furthermore, compared with the baseline model that uses the original multilingual-BERT (`m-BERT'), the success rate drops 2.2\% and the sDTW score drops 2.1\%.
This result indicates that instruction representations in one language cannot benefit from learning representation in other languages if the multilingual representation is only supervised by contrastive loss between mono-lingual instruction pairs.  

\noindent\textbf{Capturing visual concepts.} 
Our cross-lingual language representation can ground to the visual environment more easily by capturing the visual concepts in the instruction. We demonstrate that shared visual concepts in different paths are captured by our language representation. We first encode the instruction as in Eqn.~\ref{eqn:lang} with cross-lingual representation and original multilingual BERT separately. 
For every instruction, we retrieve another instruction with the highest cosine similarity under the constraints that two instructions don't correspond to the same path and equal path length. As shown in Figure~\ref{figure3}, the second row is the query instruction and the first row is its corresponding path. The following four rows correspond to the instruction-path picked with cross-lingual representation and multilingual-BERT representation. 
First, we observe in Figure~\ref{figure3} that our cross-lingual representation retrieves a Hindi instruction while the multilingual-BERT picks an English instruction. This indicates that our cross-lingual representation learns to encode instructions with similar semantics in different languages closer to each other.
Besides, we observe that in all three paths, the agent passes tables and chairs, but only in the query path and the cross-lingual paired path, the agent stops at places similar to ``bar stools". This demonstrates that the visual objects in the cross-lingual picked path are more similar to the objects in the query path.

\begin{table}
\centering
\resizebox{0.95\columnwidth}{!}{
    \begin{tabular}{cccccc}
    \hline 
       \textbf{Similarity}  &  \textbf{SR$\uparrow$} & \textbf{SPL$\uparrow$} & \textbf{NDTW$\uparrow$} & \textbf{SDTW$\uparrow$} \\ \hline
        mono & 32.9 & 30.4 & 51.4 & 28.0\\ 
        \textit{en+hi} & 32.0 & 28.3 & 48.4 & 26.9 \\ 
        \textit{en+te} & 30.0 & 27.2 & 48.8 & 25.4 \\ 
        \textit{hi+te} & 27.8 & 25.0 & 46.1 & 23.2 \\ 
        multi & 36.0 & 31.7 & 52.0 & 30.5 \\ \hline
    \end{tabular}
    }
    \caption{Comparison between language representation trained with different instruction pairs. `mono' indicates representation trained with mono-lingual instruction pairs, `multi' indicates representation trained with multilingual instruction pairs in all three languages, and \textit{`en+hi'} indicates representation trained with multilingual instruction pairs in English and Hindi only.}
    \vspace{-10pt}
    \label{table6_appendix}
\end{table}

\subsection{Effectiveness of Optimizing Similarity between Three Languages}  \label{sec:3lang}

In this section, we further show that only optimizing the similarity between a subset of languages (i.e., two out of three languages) will hurt the performance. Specifically, we train the language representation that optimizes similarity between only English and Hindi (\textit{`en+hi'}), only English and Telugu (\textit{`en+te'}), only Hindi and Telugu (\textit{`hi+te'}), and only single language (`mono'). Given paired language instructions in English and Hindi in unseen set, the average distance is 0.61 for our language representation (i.e., optimizes similarity between all three languages), 0.43 for \textit{en+te}, 1.67 for \textit{hi+te}, and 1.55 for same language only, indicating that explicitly optimizing the similarity between \textit{en+hi} helps reduce the distance between \textit{en+hi} most. Adding \textit{te} in optimization will make \textit{en+hi} farther from each other, but still much better than only optimizing \textit{hi+te}, and could also make the distance between all three languages to be closer to each other. We further show the performance of training the navigation agent with these language representations in Table~\ref{table6_appendix}. We observe that both the success rate and the nDTW score drop significantly when only training on a subset of languages. This result shows that it's crucial to train the language representation with instruction pairs in all three languages.

\begin{table*}[t]
    \centering
    \resizebox{1.8\columnwidth}{!}{
    \begin{tabular}{p{0.32\columnwidth}>{\centering\arraybackslash}p{0.08\columnwidth}>{\centering\arraybackslash}p{0.08\columnwidth}>{\centering\arraybackslash}p{0.09\columnwidth}>{\centering\arraybackslash}p{0.09\columnwidth}|>{\centering\arraybackslash}p{0.08\columnwidth}>{\centering\arraybackslash}p{0.08\columnwidth}>{\centering\arraybackslash}p{0.09\columnwidth}>{\centering\arraybackslash}p{0.09\columnwidth}|>{\centering\arraybackslash}p{0.08\columnwidth}>{\centering\arraybackslash}p{0.08\columnwidth}>{\centering\arraybackslash}p{0.09\columnwidth}>{\centering\arraybackslash}p{0.09\columnwidth}}
\hline 
\multicolumn{1}{c}{\textbf{Models}} & \multicolumn{4}{c}{\textbf{seen}} & \multicolumn{4}{c}{\textbf{unseen}} & \multicolumn{4}{c}{\textbf{$|\Delta|$}}  \\
    \hline 
    & \textbf{\small SR} &\textbf{\small SPL} & \textbf{\small NDTW} & \textbf{\small SDTW} & \textbf{\small SR} &\textbf{\small SPL} & \textbf{\small NDTW} & \textbf{\small SDTW} & \textbf{\small SR} &\textbf{\small SPL} & \textbf{\small NDTW} & \textbf{\small SDTW}\\
    \textbf{\citet{ku-etal-2020-room}} & 25.2 & - & 42.2 & 20.7 & 22.8 & - & 38.9 & 18.2 & 2.4  & - & 3.3 & 2.5 \\
    \textbf{ResNet} & 38.4 & 34.1 & 52.7 & 32.6 & 35.1 & 31.6 & 51.1 & 30.1 & 3.3 &2.5 & 1.6 & 2.5  \\
    \textbf{+visual}  & 34.1 & 31.1 & 52.7 & 28.8 & 35.6 & 32.5 & 53.7 & 30.5 & 1.5 & 1.4 & 1.0 & 1.7 \\
    \hline \\
    \end{tabular}}
    \vspace{-14pt}
    \caption{The results of adding our learned visual representation on validation seen environments and validation unseen environments. $|\Delta|$ indicates absolute performance difference between seen and unseen environments.}
    \vspace{-8pt}
    \label{table1_appendix}
\end{table*}

\subsection{Decreasing Gap between Seen and Unseen Environments} \label{sec:gap}
Most previous navigation models \cite{wang2019reinforced, ma2019selfmonitoring, majumdar2020improving} suffer from a large performance drop when moving from seen validation to unseen validation because the visual encoder overfits the low-level appearance features~\cite{ZhangTB20}. 
Our environment agnostic visual representation can decrease the performance gap between validation seen and unseen environments. As shown in Table~\ref{table1_appendix},
the nDTW gap is decreased from 1.6 to 1.0 compared with baseline model. It is also lower than the gap of a multilingual agent in \citet{ku-etal-2020-room}.

\subsection{Comparison with Other Contrastive Learning Approaches} \label{section:simcse}

In this section, we compare with SimCSE \cite{gao2021simcse}, an effective contrastive learning approach for text representation learning. We use SimCSE on our visual representation learning, where we use dropout as positives in contrastive learning. Using SimCSE to train the visual representation gets 34.8/53.0 (SR/nDTW), which is lower than our visual representation (35.6/53.7). Furthermore, we experiment with using both dropout as positives and our identified path pairs as positives. The performance decreases in nDTW score (52.4) compared with only using our identified path pairs as positives (53.7).

\begin{table}[t]
    \centering
    \resizebox{0.95\columnwidth}{!}{
    \begin{tabular}{cccccc}
    \hline 
      \textbf{Models}   &  \textbf{SR$\uparrow$} & \textbf{SPL$\uparrow$} & \textbf{NDTW$\uparrow$} & \textbf{SDTW$\uparrow$} \\ \hline
        \textbf{ResNet}  & 49.1 & 44.7 & 58.8 & 42.0 \\
         \textbf{+text} & 49.0 & 45.2 & 59.5 & 42.3 \\
         \textbf{+visual} & 50.4 & 46.3 & 60.3 & 43.4 \\ 
    \textbf{CLEAR} & \textbf{50.5} & \textbf{46.4} & \textbf{60.6} & \textbf{43.3} \\  \hline
    \textbf{ResNet-zero} & 30.9 & 27.9& 49.0 & 26.3 \\ 
    \textbf{CLEAR-zero}&  \textbf{35.4} & \textbf{30.1} & \textbf{49.0} & \textbf{28.2} \\ \hline
    \end{tabular}
    }
    \vspace{-3pt}
    \caption{Results on R2R validation unseen environments. ``CLEAR" (based on ResNet) transfers the language and visual representation from RxR dataset, and ``ResNet" is the baseline model that uses multilingual BERT and pre-trained ResNet. ``ResNet-zero" and ``CLEAR-zero" are zero-shot performance of baseline and our approach on R2R dataset.}
    \vspace{-14pt}
    \label{table7}
\end{table}

\subsection{Generalization to Other VLN Tasks} \label{sec:generalization}
We further evaluate our CLEAR approach's generalizability on Room-to-Room (R2R) dataset \cite{anderson2018vision} and Cooperative Vision-and-Dialog Navigation (CVDN) dataset \cite{thomason:arxiv19}, in which we directly transfer our CLEAR approach and train on the navigation task on R2R and CVDN. R2R and CVDN follow the same training, validation seen, and validation unseen split of environments as Room-Across-Room dataset. The main difference is that the language instructions in R2R and CVDN is monolingual (i.e., English). Besides, instructions in CVDN are multi-round dialogues between the navigator and the oracle.
Our baseline model uses multilingual BERT to encode instructions and the ResNet pretrained on ImageNet to extract image features. The cross-lingual language representation and environment-agnostic visual representation is trained on RxR dataset (as in Sec.~\ref{section:language} and Sec.~\ref{section:visual}). We then train the navigation agent on R2R dataset and CVDN dataset with the language and visual encoder initialized from our CLEAR representation. 

As shown in Table~\ref{table7}, on R2R dataset, our learned representation outperforms the baseline by 1.4\% in success rate and 1.8\% in nDTW. Furthermore, we show that the zero-shot performance of our approach improves the baseline by 4.5\% in success rate and 2.2\% in SPL on R2R dataset. On CVDN dataset, our learned representation outperforms the baseline by 0.74 in Goal Progress (4.05 vs. 3.31) after training on CVDN dataset, and outperforms the baseline by 0.42 in Goal Progress (0.92 vs. 0.50) in the zero-shot setting. Goal Progress measures the progress made towards the target location and is the main evaluation metric in CVDN. 
This result demonstrates that our learned cross-lingual and environment agnostic representation could generalize to other tasks.

\subsection{Generalization to Other VLN Agents} \label{sec:vln_agent}

We further evaluate our CLEAR approach's generalizability to another VLN agent. Specifically, we adapt CLEAR to SotA VLN agent HAMT \cite{chen2021history}. With the pre-trained weights released in HAMT, we further learn the text representation and visual representation with our approach. Adapting CLEAR to HAMT achieves 57.2\% in success rate and 65.6\% in nDTW score, which is 0.7\% higher than HAMT in success rate and 2.5\% higher than HAMT in nDTW score on RxR validation unseen set, demonstrating the effectiveness of our proposed approach over SotA VLN models.

\section{Conclusion}
In this paper, we presented the CLEAR method that learns a cross-lingual and environment-agnostic representation. We demonstrated that our cross-lingual language representation captures more visual semantics and our environment-agnostic representation generalizes better to unseen environments. Our experiments on Room-Across-Room dataset suggest that our CLEAR method improved the performance in all evaluation metrics over a strong baseline.
Furthermore, we qualitatively and quantitatively analyze the effectiveness of every component of our CLEAR approach and its generalizability to other tasks and base VLN agents.

\section*{Ethics Statement}
In this paper, we presented a method to learn cross-lingual and environment-agnostic representations for Vision-and-Language Navigation. Vision-and-Language Navigation task can be used in many real-world applications, for example, a home service robot can bring things to the owner based on natural language instructions. Our learned representations enable the agent to understand multilingual instructions and improve agents' generalizability to unseen environments. 
However, currently we learn our cross-lingual representation from three languages (i.e., English, Hindi, and Telugu) due to dataset availability, which might limit its generalization to other languages. Besides, similar to other instruction-following agents, our agent might fail to reach the target given some instructions, which requires further human assistance.

\section*{Acknowledgement}
We thank the reviewers for their helpful comments. This work was supported by ARO W911NF2110220, ONR N000141812871, DARPA KAIROS FA8750-19-2-1004, Google Focused Award. The views contained in this article are those of the authors and not of the funding agency.

\bibliography{custom}
\bibliographystyle{acl_natbib}

\appendix

\section*{Appendix}

\section{Overview}
In this supplementary, we provide a detailed description of our navigation model structure (Sec.~\ref{sec:navigation_model_appendix}), representation learning and navigation learning objective (Sec.~\ref{sec:learning}), dataset (Sec.~\ref{sec:dataset}), evaluation metrics (Sec.~\ref{sec:eval}), implementation details (Sec.~\ref{sec:implementation}), and additional analysis in the last four sections. In this analysis, we first show that using object-matching as constraints during visual representation learning improves the nDTW score (Sec.~\ref{sec:object-matching}). 
Then we show that our CLEAR approach decreases the performance variance among different environments (Sec.~\ref{sec:std}) and learn better alignment between the instruction and the environment (Sec.~\ref{sec:alignment}). We further analyze whether the word representation from our learned cross-lingual representation also learn the visual/spatial information (Sec.~\ref{sec:word representation}). Moreover, we investigate the effect of filtering out low-quality paths (Sec.~\ref{section:filter}). Lastly, we show the high correspondence between instruction similarity and path pair alignment in Sec.~\ref{section:similarity}.

\section{Navigation Model} \label{sec:navigation_model_appendix}
Our navigation agent follows the decoder structure as \citet{tan-etal-2019-learning}. 

At each time step $t$, the agent perceives a panoramic view of the current location. The panoramic view is discretized into 36 single views $\{o_{t,m}\}_{m=1}^{36}$ (12 angles and 3 camera poses per angle). Given the visual representation for each view $\widehat{v}_{t,m}$, we concatenate it with the orientation feature to get the view features $\{f_{t,m}\}_{m=1}^{36}$: 
\begin{align}
\label{eqn:f}
    f_{t,m} = [\widehat{v}_{t,m}; (\cos\theta_{t,m}, \sin\theta_{t,m}, \cos\phi_{t,m}, \sin\phi_{t,m})]
\end{align}
where $\theta_{t,m}$ and $\phi_{t,m}$ the heading and elevation of view $o_{t,m}$.

As a reaction to the input, the agent needs to select one of the $K$ navigable locations as an action. The action is represented as the orientation features (heading and elevation) between the current viewpoint and the chosen navigable viewpoint. The navigation decoder takes the attended visual feature $\widehat{f}_{t}$ of the current viewpoint and the previous action embedding $a_{t-1}$ as input, and updates its environment-aware context vector $h_t$:
\begin{align}
    \gamma_{t,m} &= \mathrm{Softmax}_m(f_{t,m}^TW_f\widehat{h}_{t-1}) \\ 
    \widehat{f}_{t} &= \sum_{m}{\gamma_{t,m}f_{t,m}} \\
    h_t &= \mathrm{LSTM}([\widehat{f}_t; a_{t-1}], \widehat{h}_{t-1})
\end{align}
where $a_{t-1}$ is represented as the orientation features $(\cos\theta_{t-1,k^\star}, \sin\theta_{t-1,k^\star}, \cos\phi_{t-1,k^\star}, \sin\phi_{t-1,k^\star})$ of the chosen navigable viewpoint $k^\star$ at time step $t-1$, and $\widehat{h}_{t-1}$ is the instruction-aware context vector that incorporates the attended instruction information. The navigator calculates the probability of moving to the $k\mbox{-}th$ navigable location based on the alignment between the visual feature $g_{t,k}$ of that navigable location and the instruction-aware context vector $\widehat{h}_t$:
\begin{align}
    \rho_{t,j} &= \mathrm{Softmax}_j(\widehat{w}_j^TW_lh_t) \\
    u_t &= \sum_j{\rho_{t,j}\widehat{w}_j} \\
    \widehat{h}_t &= \mathrm{tanh}(W_m[u_t; h_t]) \\
    p(a_t = k) &= \mathrm{Softmax}_k(g_{t,k}^TW_a\widehat{h}_t)
\end{align}
where $g_{t,k}$ is constructed similarly as $f_{t,i}$ in Eqn.~\ref{eqn:f}, and $\widehat{w}_j$ is the language representation.

\section{Learning} \label{sec:learning}
Our CLEAR agent has two stages of learning: representation learning and navigation learning. 

In the representation learning stage, given a pair of instructions that correspond to the same navigation path, we train the shared multilingual encoder to generate representations of paired instructions close to each other by optimizing a contrastive loss $L_{lang}$. Furthermore, we train the visual encoder to learn the connections between paths with similar instructions by optimizing the contrastive loss $L_{visual}$. The representation learning process transfers the language representation to domain-specific language representation and adapts the visual representation to learn the correlation underlying the navigation environments. 

In the navigation learning stage, we use a mixture of imitation learning and reinforcement learning to train the agent on the navigation task as in \citet{tan-etal-2019-learning}. 

In imitation learning, we use teacher-forcing to determine the next navigable viewpoint. Different from previous methods \cite{hong2020recurrent, tan-etal-2019-learning, huang2019transferable} that takes the shortest path as the teacher action, our teacher action $a_t^{\star}$ at each time step $t$ is picked based on the given ground truth path between the start point and target point. The agent tries to imitate the teacher action $a_t^\star$ by minimizing the negative log probability:
\begin{align}
    L_{IL} &= \sum_t{-{a_t^\star}\log p_t}
\end{align}

We combine reinforcement learning with imitation learning to learn a more generalizable agent. At each time step $t$, the agent samples an action $a_t$ from the predicted distribution $p_t(a_t)$. We follow \cite{hong2020recurrent} to do the reward shaping. The immediate reward at each time step $t$ consists of three parts. First, if the agent moves closer to the target viewpoint, a positive reward +1 is given, otherwise the agent receives a negative reward -1. Second, to encourage instruction following, we include normalized Dynamic Time Warping (nDTW) score in the reward. The agent gets a positive reward if the nDTW score for the navigated path increases. Lastly, the agent receives a negative reward if it misses the target. When the agent predicts the ``STOP" action, the agent will receive a +3/-3 reward based on whether the agent is within 3 meters from the target viewpoint. We use Advantage Actor-Critic \cite{mnih2016asynchronous} to train the agent. 

The navigation loss $L_{nav}$ is a weighted combination of imitation learning loss and reinforcement learning loss.
\begin{align}
    L_{nav} &= L_{RL} + \lambda L_{IL} 
\end{align}

\section{Dataset} \label{sec:dataset}
We evaluate our agent on the Room-Across-Room (RxR) dataset \cite{ku-etal-2020-room}. The dataset is built on the Matterport3D simulator \cite{anderson2018vision}. It contains 126,069 human-annotated instructions with an average instruction length of 78. The dataset is split into training set, seen validation set, unseen validation set, and test set. In the unseen validation set and test set, the environments do not appear in the training set. Thus the performance on these two sets show the model's generalizability to new environments. There are 16,522 paths in total, and each path is annotated in 3 languages (and 3 instructions per language on average). The training set contains 11,089 paths, the seen validation set contains 1,232 paths, the unseen validation contains 1,517 paths, and the test set contains 2,684 paths.

\section{Evaluation Metrics}  \label{sec:eval}
To evaluate the performance of our model, we follow the metrics used in the Room-Across-Room paper \cite{ku-etal-2020-room}. The metrics include: (1) Success Rate (SR): We consider a success for navigation if the agent stops less than 3m from the target location. (2) Success rate weighted by Path Length (SPL) \cite{anderson2018evaluation}: This metric penalizes the navigation with long paths (i.e., when both navigations reach the target, the navigation with shorter path length has a higher SPL score). (3) normalized Dynamic Time Warping (nDTW) \cite{magalhaes2019effective}: This metric measures the path fidelity by penalizing deviations from the reference path. The agent navigates to the target through the shortest path instead of instruction following will be penalized. (4) success rate weighted by Dynamic Time Warping (sDTW) \cite{magalhaes2019effective}: This metric only considers nDTW of successful navigation and ignores failed navigation. Normalized Dynamic Time Warping (nDTW) is the main metrics for RxR and Success Rate (SR) and Success rate weighted by Path Length (SPL) are the main metrics for R2R.

\begin{table}
\centering
    \begin{tabular}{cccccc}
    \hline 
       \textbf{Methods}  &  \textbf{SR$\uparrow$} & \textbf{SPL$\uparrow$} & \textbf{NDTW$\uparrow$} & \textbf{SDTW$\uparrow$} \\ \hline
        \textbf{+visual}  & 35.6 & 32.5 & 53.7 & 30.5 \\
        \textbf{-sample} & 37.8 & 33.7 & 53.0 & 32.1 \\
    \textbf{-object}   & 36.6 & 33.0 & 52.4 & 30.9 \\  \hline 
    \end{tabular}
    \caption{Comparison between visual representation trained with objects constraints (`+visual'), without sampling strategy (`-sample') and without object constraints (`-object') on validation unseen sets. nDTW is the main metric for Room-Across-Room (RxR) dataset.}
    \label{table3_appendix}
\end{table}

\begin{table*}[ht]
\begin{small}
    \centering
    \begin{tabular}{p{0.3\columnwidth}>{\centering\arraybackslash}p{0.1\columnwidth}>{\centering\arraybackslash}p{0.1\columnwidth}>{\centering\arraybackslash}p{0.1\columnwidth}>{\centering\arraybackslash}p{0.12\columnwidth}>{\centering\arraybackslash}p{0.12\columnwidth}|>{\centering\arraybackslash}p{0.1\columnwidth}>{\centering\arraybackslash}p{0.1\columnwidth}>{\centering\arraybackslash}p{0.12\columnwidth}>{\centering\arraybackslash}p{0.12\columnwidth}}
\hline 
\multicolumn{1}{c}{\textbf{Environment}} & \multicolumn{1}{c}{\textbf{\# Data}}& \multicolumn{4}{c}{\textbf{ResNet}} & \multicolumn{4}{c}{\textbf{CLEAR}}  \\
    \hline 
    &  &\textbf{SR} &\textbf{SPL} & \textbf{NDTW} & \textbf{SDTW} & \textbf{SR} &\textbf{SPL} & \textbf{NDTW} & \textbf{SDTW} \\
    \begin{minipage}{.3\columnwidth}
      \includegraphics[width=\columnwidth]{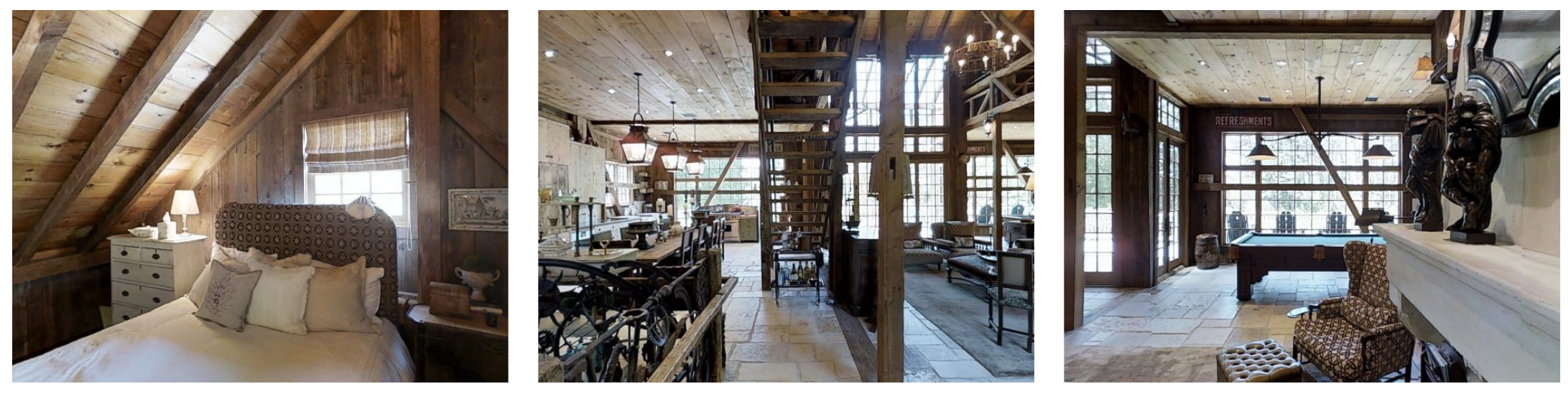}
    \end{minipage} & 1206 &32.4 & 26.7 & 49.8 & 26.5 & 35.9 & 30.2 & 50.0 & 28.0\\ 
    \begin{minipage}{.3\columnwidth}
      \includegraphics[width=\columnwidth]{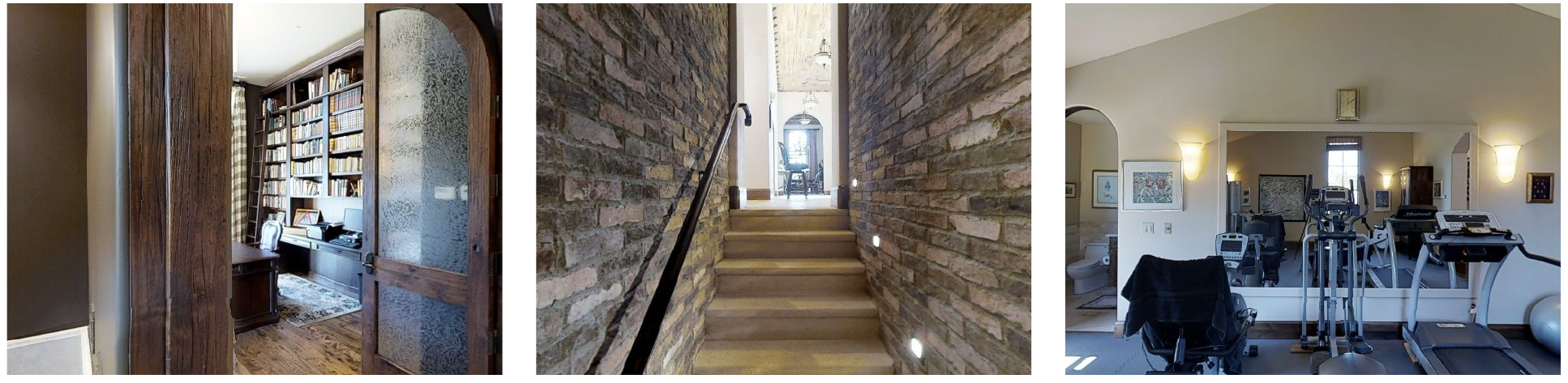}
    \end{minipage} & 2177 &27.0 & 23.8 & 41.3 & 22.3 & 28.6 & 26.0 & 47.5 & 24.4\\ 
    \begin{minipage}{.3\columnwidth}
      \includegraphics[width=\columnwidth]{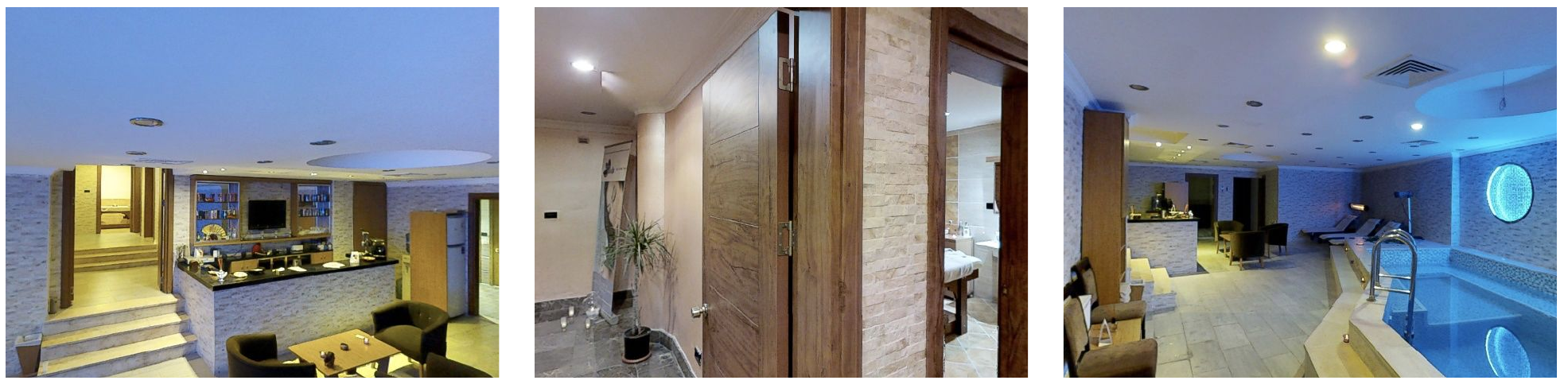}
    \end{minipage} & 567 &38.1 & 34.3 & 56.9 & 31.8 & 48.0 & 44.8 & 64.4 & 40.4 \\
    \begin{minipage}{.3\columnwidth}
      \includegraphics[width=\columnwidth]{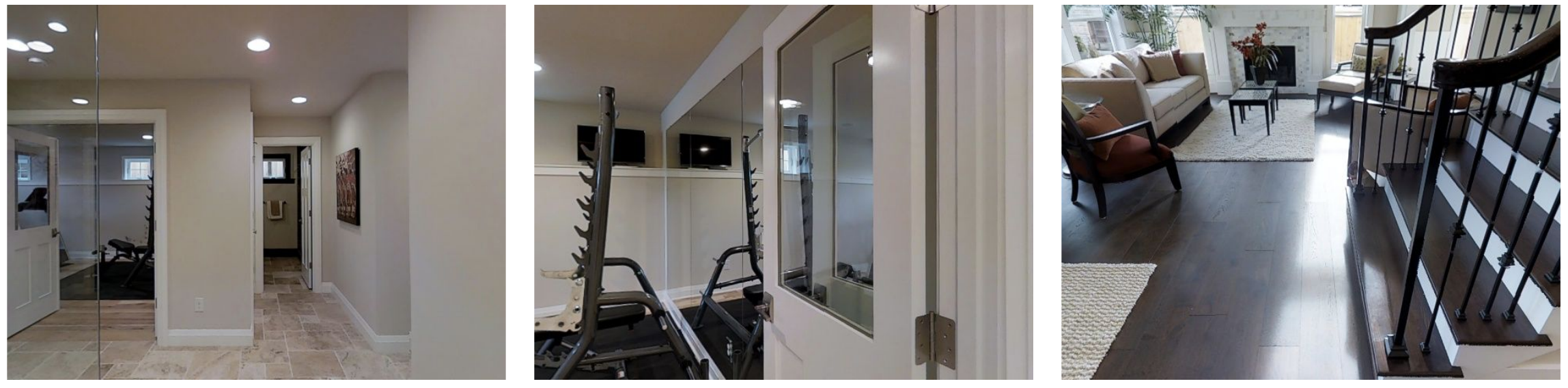}
    \end{minipage} & 1692 &38.3 & 34.8 & 56.1 & 33.2 & 39.5 & 36.0 & 57.6 & 33.4 \\
    \begin{minipage}{.3\columnwidth}
      \includegraphics[width=\columnwidth]{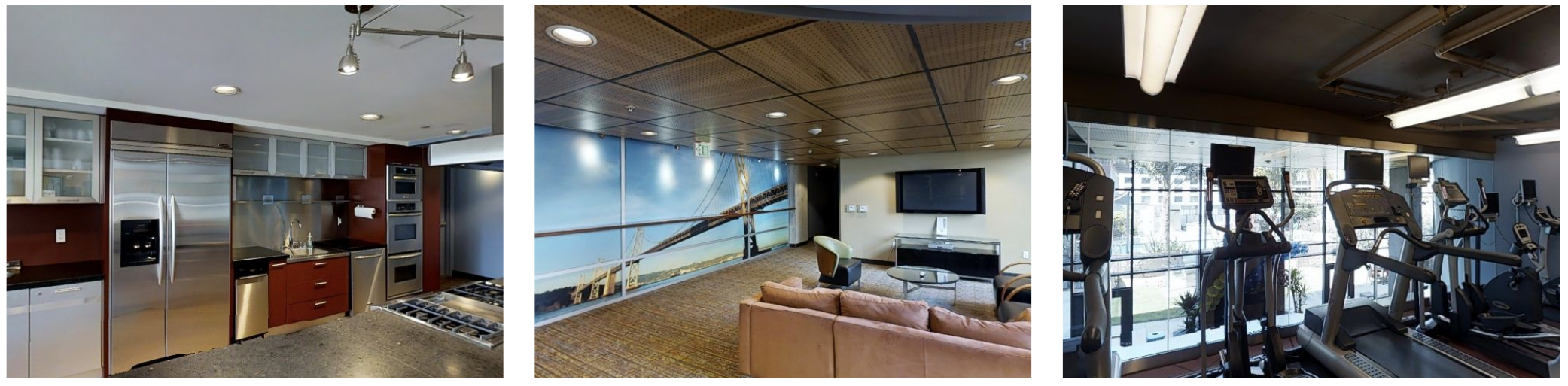}
    \end{minipage} &153 &57.5 & 54.7 & 72.1 & 53.1 & 64.1 & 60.0 & 74.5 & 57.3 \\
    \begin{minipage}{.3\columnwidth}
      \includegraphics[width=\columnwidth]{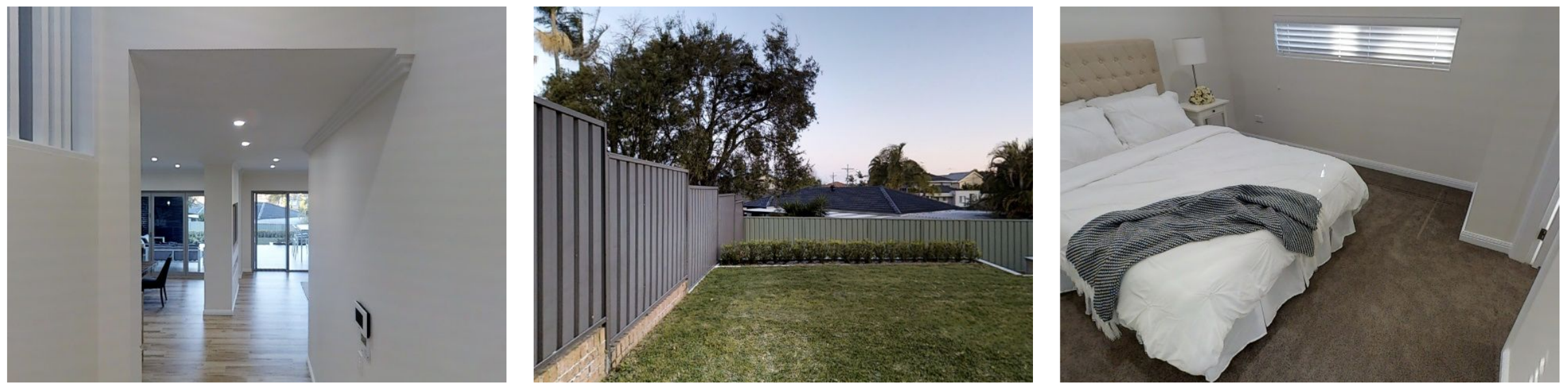}
    \end{minipage} & 1404 &42.7 & 38.7 & 58.3 & 37.8 & 41.3 & 39.1 & 61.1 & 36.6 \\ 
    \begin{minipage}{.3\columnwidth}
      \includegraphics[width=\columnwidth]{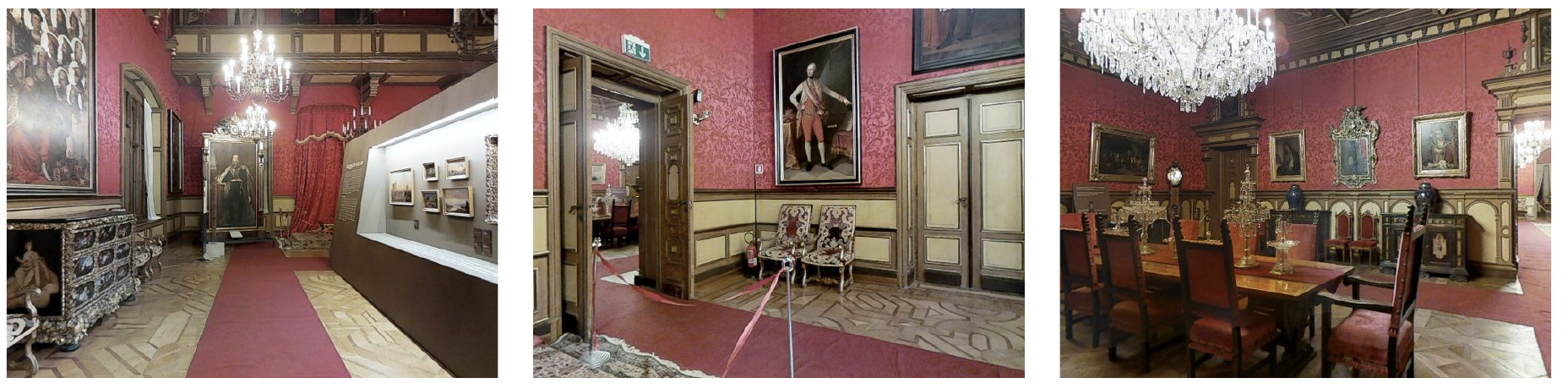}
    \end{minipage} & 900 &52.0 & 49.5 & 67.9 & 46.3 & 45.7 & 44.2 & 65.9 & 41.0 \\
    \begin{minipage}{.3\columnwidth}
      \includegraphics[width=\columnwidth]{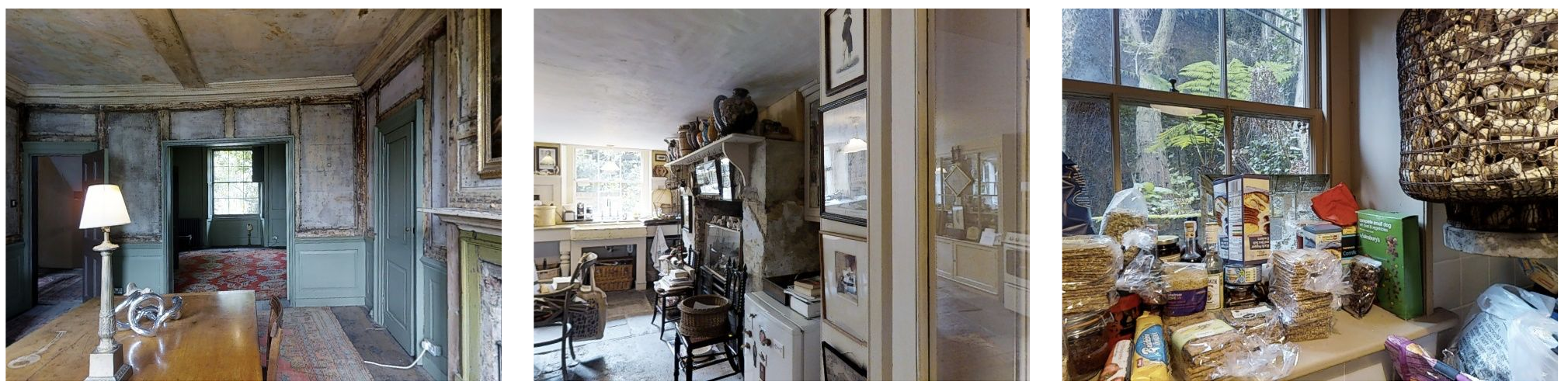}
    \end{minipage} &2223 &40.8 & 36.9 & 57.7 & 35.5 & 44.0 & 39.2 & 60.3 & 38.2 \\
    \begin{minipage}{.3\columnwidth}
      \includegraphics[width=\columnwidth]{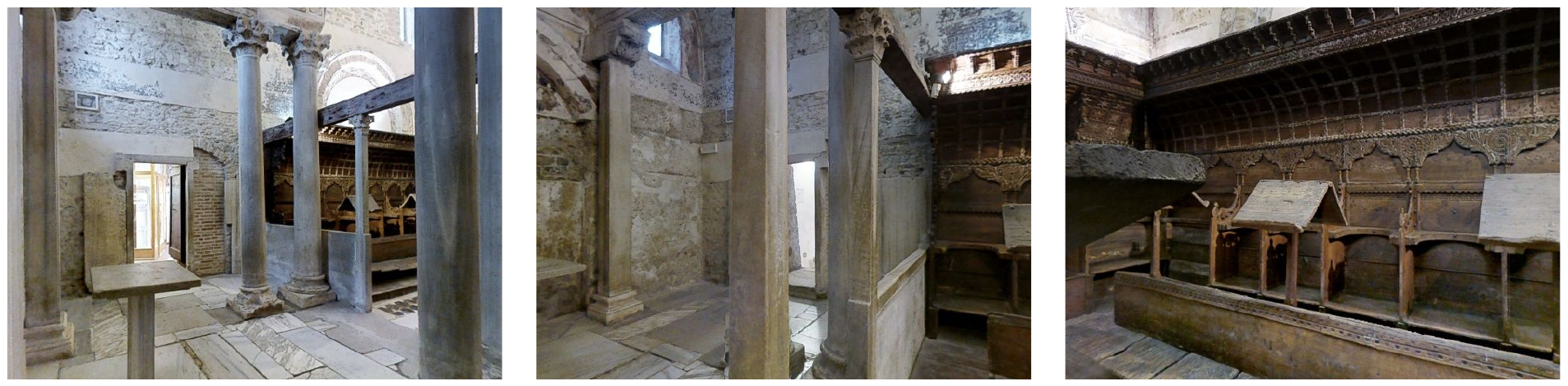}
    \end{minipage} & 18 & 38.9 & 32.7 & 59.4 & 34.0 & 50.0 & 45.1 & 70.8 & 47.1 \\
    \begin{minipage}{.3\columnwidth}
      \includegraphics[width=\columnwidth]{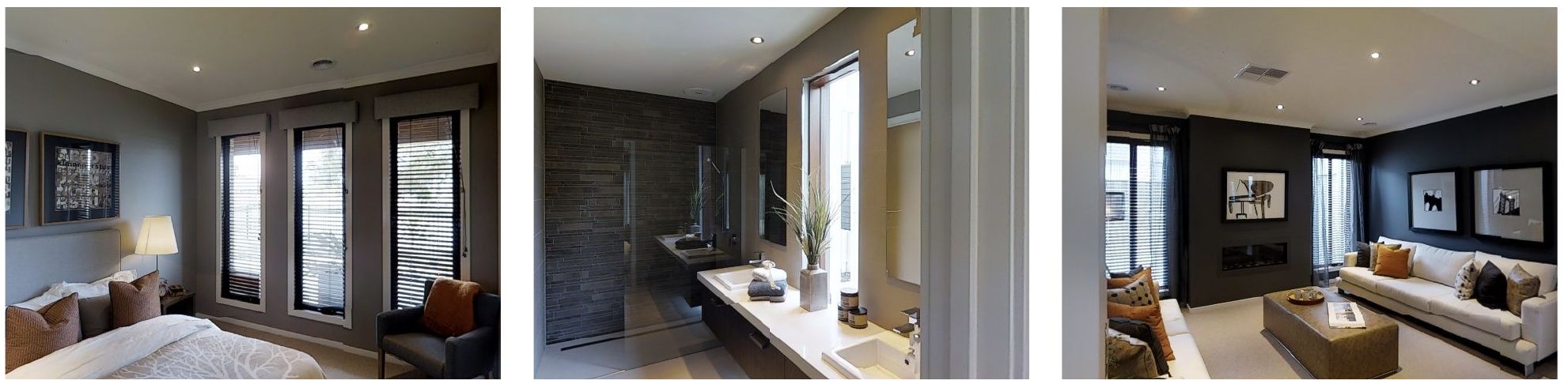}
    \end{minipage} & 1152&42.4 & 38.6 & 54.6 & 36.1 & 38.1 & 35.2 & 55.1 & 33.3 \\ 
    \begin{minipage}{.3\columnwidth}
      \includegraphics[width=\columnwidth]{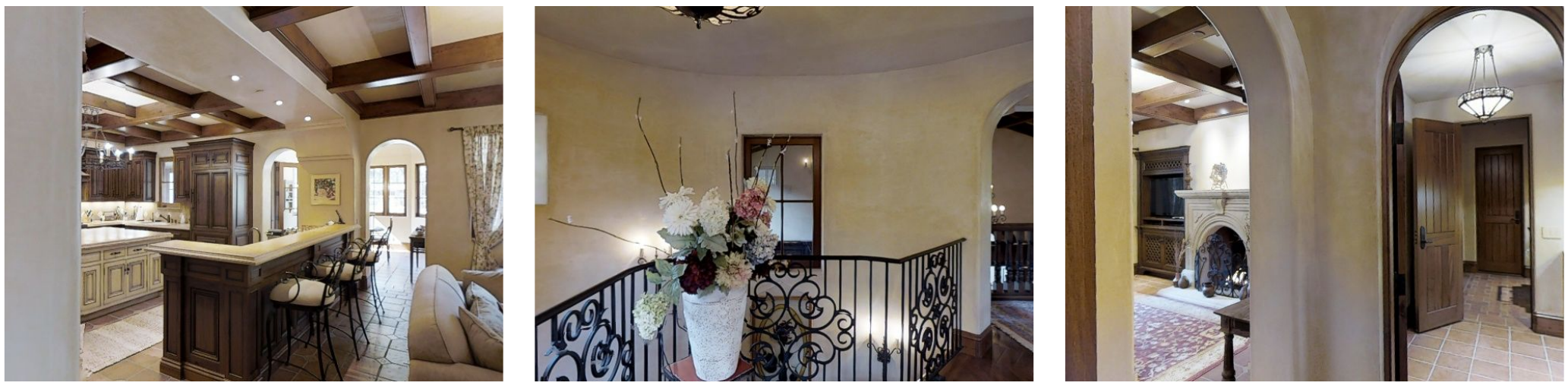}
    \end{minipage} & 2160 &18.1 & 16.4 & 34.6 & 14.7 & 15.9 & 13.1 & 37.1 & 13.1 \\
    \hline \\
    \end{tabular}
    \vspace{-14pt}
    \caption{The results of our CLEAR method and ResNet baseline on different environments in validation unseen set. \# Data means the number of instruction-path pairs for each environment.}
    \label{table2_appendix}
    \end{small}
\end{table*}

\section{Implementation Details} \label{sec:implementation}
In our experiments, we learn the shared multilingual representation based on cased multilingual $\text{BERT}_\text{BASE}$. The instruction is truncated from the end with a maximum sequence length of 160. For the pre-trained vision model, we compare performance between image features extracted from ImageNet-pre-trained \cite{russakovsky2015imagenet} ResNet-152 \cite{he2016deep} and CLIP-pre-trained \cite{radford2021learning} vision transformer (ViT-B/32) \cite{dosovitskiy2010image} (abbreviated as `CLIP feature' later). The 27 object classes are: `drawer', `faucet', `cabinet', `hinge', `cushion', `sofa', `chair', `pillow', `armchair', `lamp', `vase', `knob', `curtain', `statue(sculpture)', `doorknob', `vent', `lightbulb', `flowerpot', `book', `pipe', `painting', `wall socket', `bed', `mirror', `television set', `flower arrangement', `chandelier'. The navigation decoder's hidden size is 768 and the action embedding size is 128. The language encoder is optimized with AdamW \cite{loshchilov2017decoupled} with linear-decayed learning rate. The peak learning rate is 4e-5 for both the representation learning and the navigation agent learning stage. The visual encoder, the navigation decoder, and the discriminator are optimized with RMSProp \cite{hinton2012neural} with learning rate 1e-4. The weight $\lambda$ we use to combine loss is set to be 0.4 for the ResNet-based full model and 0.2 for the CLIP-based full model. The batch size for training ResNet features and CLIP features are 12 and 16, respectively. During training, CLIP model is around 1.5 times faster than ResNet model in this setting since CLIP features are 512 dimensions while ResNet features are 2048 dimensions. To keep roughly the same amount of training time, we train the agent with ResNet features for 100K iterations, while we train model CLIP-ViT features for 150K iterations.

\section{Analysis: Effectiveness of Object-Matching Constraints}\label{sec:object-matching}
Our visual representation learning optimizes the similarity between panoramic views at each step of the semantically-aligned path pairs. Since paths are not fully-aligned, we use object-matching as a constraints to filter out panoramic view pairs that don't contain same objects. As shown in Table~\ref{table3_appendix}, the visual representation trained with fixed object classes as constraints (`-sample') improve the nDTW score (the main metric for RxR dataset) by 0.6\% compared with the visual representation trained without object-matching constraints (`-object'), suggesting that using object-matching as constraints help learn a better visual representation. Besides, the sampling strategy (i.e., randomly sample 10 object classes from 27 object classes during each iteration) also helps the visual representation learning (`+visual'), further improving the nDTW score by 0.7\% compared with the visual representation learned with fixed 27 object classes (`-sample'). In total, our object-matching constraints and sampling strategy (`+visual') improves the performance by 1.3\% in nDTW score compared with learning without object constraints (`-object').

\begin{figure*}[t]
\begin{center}
\includegraphics[width=0.9\linewidth]{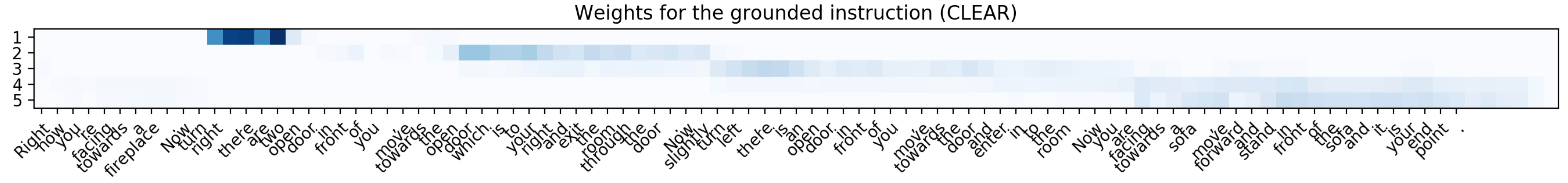}
\includegraphics[width=0.9\linewidth]{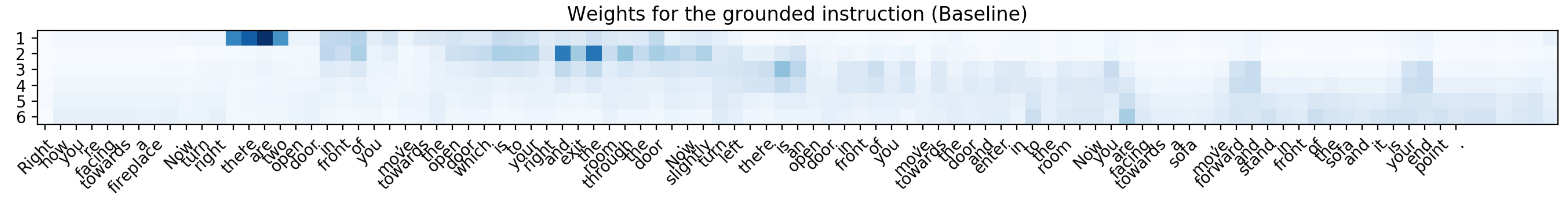}
\includegraphics[width=0.9\linewidth]{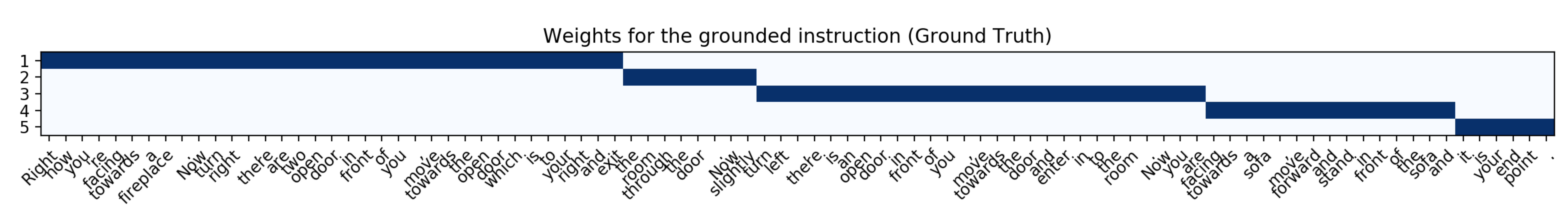}
\end{center}
\vspace{-12pt}
  \caption{The attention weights for the grounded instruction for our CLEAR model, ResNet based baseline model, and ground truth from RxR dataset.}
\label{figure1_appendix}
\end{figure*}

\section{Analysis: Performance Variance Reduction among Different Environments} \label{sec:std}

We demonstrate that our CLEAR approach could decrease the performance variance (i.e., performance's standard deviation) among different environments. Intuitively, we hope the agent to perform equally well in different environments instead of getting high performance by only learning to navigate through several easy environments. We show the results for 11 environments in validation unseen set in Table \ref{table2_appendix}. Our CLEAR approach (`+both' as in Table 2 in the main paper) outperforms the baseline model (`ResNet' as in Table~\ref{table2} in the main paper) in most of the environments. Moreover, the weighted standard deviation (weighted by \# Data in Table~\ref{table2_appendix}) of our CLEAR approach is lower than the baseline model. Specifically, the standard deviation of nDTW score for our CLEAR approach is 9.24 while the standard deviation of nDTW score for the baseline model is 10.01, suggesting that our CLEAR approach decreases the performance variance between different environments.

\begin{table}[t]
\begin{small}
    \centering
    \begin{tabular}{p{0.15\columnwidth}p{0.7\columnwidth}}
    \hline 
      \textbf{Word}   &  \textbf{Top-5} \\ \hline
      \multirow{2}*{\textbf{kitchen}} & `island', `counter', `maker', `din', `\#\#iding'\\ \cline{2-2}
      ~ & `living', `counter', `room', `table', `house' \\ \hline
      \multirow{2}*{\textbf{fire}} & `\#\#place', `over', `place', `chair', `\#\#fas' \\ \cline{2-2}
      ~ &  `display',  `study', `family', `living', `coffee'\\
    \hline
    \end{tabular}
    \caption{Top-5 closest tokens for `Kitchen' and `fire'. Top-row: tokens picked by our cross-lingual representation. Bottom-row: tokens picked by multilingual BERT baseline. }
    \label{appendix_table4}
\end{small}
\end{table}

\section{Analysis: Word Representation from Cross-Lingual Model}\label{sec:word representation}
The visual semantics are injected during learning the cross-lingual language representation by maximizing the similarity between full instruction sentences (representation of `CLS' token). However, it's unclear that whether the word-level representation also learned such visual information. In this section, we investigate whether the learning encodes spatially close words/objects closer to each other. As shown in Table~\ref{appendix_table4}, we check the top-5 close words to `kitchen', and `fire' from a vocabulary of 2754 English tokens. We see that our cross-lingual representation puts words that appear spatially near each other close (e.g. `kitchen' and `island'/`dinning', `fire' and `chair'/`fireplace') while m-BERT representation fails (e.g. `kitchen' and `room'/`house', `fire' and `family'/`study').

\section{Analysis: Alignment between Instructions and Environments} \label{sec:alignment}
The Room-Across-Room dataset provides ground-truth alignment between instructions and navigation paths. To demonstrate that our CLEAR approach learns a good alignment between instructions and paths, we not only compare our CLEAR approach with the baseline approach, but also compare it with the ground truth alignment provided in the RxR dataset. The attention weights for grounded instruction for CLEAR, Baseline, and Ground Truth are shown in Figure \ref{figure1_appendix}. We observe that our CLEAR model successfully attends to sub-instructions ``turn right", ``move towards the open door to your right and exit the room through the door", ``slightly turn left", ``move towards and stand in front of the sofa" sequentially. Although the baseline model also successfully executes the first two sub-instructions ``turn right" and ``move towards the open door", yet the baseline agent gets lost in the later navigation. Furthermore, the alignment learned by our CLEAR approach matches better with the ground truth alignment provided in the RxR dataset.

\begin{table}
\begin{small}
\centering
    \begin{tabular}{cccccc}
    \hline 
       \textbf{Similarity}  &  \textbf{SR$\uparrow$} & \textbf{SPL$\uparrow$} & \textbf{NDTW$\uparrow$} & \textbf{SDTW$\uparrow$} \\ \hline
        0.00 & 35.6 & 32.5 & 53.7 & 30.5 \\
        0.90 & 36.2 & 32.2 & 51.9 & 30.7 \\
    0.95 & 38.6 & 34.3 & 53.3 & 33.0 \\ 
    0.98 & 38.6 & 34.3 & 52.9 & 32.9 \\ 
    0.99 & 37.8 & 33.5 & 52.6 & 32.0 \\
    1.00 & 30.9 & 28.0 & 49.7 & 26.1 \\ \hline
    \end{tabular}
    \caption{Performance in validation unseen environment when filtering out different percentages of data in training our visual representation. 0.90 means filter out data with similarity score less than 0.90. }
    \label{table5_appendix}
\end{small}
\end{table}

\section{Analysis: Filtering out Low Quality Path Pairs} \label{section:filter}
We investigate whether filtering out low-quality path pairs during visual representation learning could further improve the performance. Since our identified path pairs are retrieved based on the similarity between instructions, we hypothesize that the path pair is aligned better if having a higher instruction similarity score. 
Thus, we experiment with filtering out instruction pairs that have a cosine similarity score less than 0.90, 0.95, 0.98, and 0.99, and then train the visual representation with filtered data and object-matching constraints. 
The proportion of filtered-out data is 1\%, 6\%, 28\% and 58\% respectively. 
We also experiment with filtering out 0\% and 100\%.
Filtering out 0\% of the data is the same to our proposed environment-agnostic visual representation (`+visual' in Table~\ref{table2}) and filtering out 100\% of the data is analogous to randomly initialize the visual encoder\footnote{Note that filtering out 100\% of the data is not the same as the baseline model (`ResNet' in Table~\ref{table2}). 
The baseline model does not have the visual encoder we introduced in Sec.~\ref{section:visual}}. 
We then train our environment-agnostic representation (in Sec.~\ref{section:visual}) based on the remaining data and show its performance on the validation unseen environments. 
As shown in Table~\ref{table5_appendix}, though the success rate improves when filtering out some path pairs with lower quality, not filtering out any path pairs achieve the highest nDTW score. This demonstrates that using object-matching constraints without filtering out path pairs with low instruction similarity is enough for learning a good visual representation. Furthermore, we see a significant performance drop when not fine-tuning the visual representation on any data, which indicates that training the visual encoder with semantically-aligned path pairs is important for agent performance.

\section{Analysis: Correspondence between Instruction Similarity and Path Pair Alignment} \label{section:similarity}

In this section, we show that instruction pairs that have high similarity have similar BLEU score and ROUGE score to the instruction pairs that corresponding to the same path. Specifically, the BLEU-1 and ROUGE-L score for instruction pairs that have high similarity are 0.42 and 0.320, and the BLEU-1 and ROUGE-L score for the instruction pairs that corresponding to the same path are 0.41 and 0.323. Randomly picking gets 0.37 BLEU-1 score and 0.295 ROUGE-L score. These results indicate that high similarity instruction pairs may be of competitive quality as the instruction pairs that corresponding to the same path, and can be used to pick the semantically-aligned path pairs.

\end{document}